\documentclass[10pt,twocolumn,letterpaper]{article}

\usepackage{authblk}

\usepackage[pagenumbers]{cvpr}
\usepackage{multirow}
\usepackage{fancyhdr}
\usepackage{array}
\usepackage{bbm}
\usepackage{graphicx}
\usepackage{amsmath}
\usepackage{amssymb}
\usepackage{booktabs}
\usepackage{bbding}
\usepackage{color}
\usepackage{appendix}
\usepackage{subcaption}
\usepackage{graphicx}
\usepackage[utf8]{inputenc}
\usepackage{CJKutf8}

\definecolor{cvprblue}{rgb}{0.21,0.49,0.74}
\usepackage[pagebackref,breaklinks,colorlinks,citecolor=cvprblue]{hyperref}
\usepackage{xcolor}
\definecolor{DarkGreen}{rgb}{0.43, 0.68, 0.28}

\newcommand\keywords[1]{\textbf{Keywords}: #1}

\title{NOVA-3D: Non-overlapped Views for 3D Anime Character Reconstruction}

\author[1,2]{Hongsheng Wang \textsuperscript{*}}
\author[2]{Nanjie Yao \textsuperscript{*}}
\author[2]{Xinrui Zhou}
\author[1]{Shengyu Zhang}
\author[3]{Huahao Xu \textsuperscript{$\dag$}}
\author[1]{\\ Fei Wu}
\author[2]{Feng Lin}

\affil[1]{Zhejiang University, China}
\affil[2]{Zhejiang Lab, China}
\affil[3]{Gameday Inc., China}

\makeatletter
\renewcommand\AB@affilsepx{, \protect\Affilfont}
\makeatother

\begin{document}

\maketitle

\thispagestyle{fancy}

\lfoot{This work has been submitted to the IEEE for possible publication. Copyright may be transferred without notice, after which this version may no longer be accessible.}

% \cfoot{}

\renewcommand{\headrulewidth}{0mm}

\begin{CJK}{UTF8}{gbsn}
\renewcommand{\thefootnote}{\fnsymbol{footnote}}
\footnotetext[1]{These authors contributed equally to this work.}
\footnotetext[2]{Corresponding Author.}
\begin{abstract}

In the animation industry, 3D modelers typically rely on front and back non-overlapped concept designs to guide the 3D modeling of anime characters. However, there is currently a lack of automated approaches for generating anime characters directly from these 2D designs. In light of this, we explore a novel task of reconstructing anime characters from non-overlapped views. This presents two main challenges: existing multi-view approaches cannot be directly applied due to the absence of overlapping regions, and there is a scarcity of full-body anime character data and standard benchmarks. To bridge the gap, we present \textbf{N}on-\textbf{O}verlapped \textbf{V}iews for 3D \textbf{A}nime Character Reconstruction (NOVA-3D), a new framework that implements a method for view-aware feature fusion to learn 3D-consistent features effectively and synthesizes full-body anime characters from non-overlapped front and back views directly. To facilitate this line of research, we collected the NOVA-Human dataset, which comprises multi-view images and accurate camera parameters for 3D anime characters. Extensive experiments demonstrate that the proposed method outperforms baseline approaches, achieving superior reconstruction of anime characters with exceptional detail fidelity. In addition, to further verify the effectiveness of our method, we applied it to the animation head reconstruction task and improved the state-of-the-art baseline to 94.453 in SSIM, 7.726 in LPIPS, and 19.575 in PSNR on average. Codes and datasets are available at \url{https://wanghongsheng01.github.io/NOVA-3D/}.

\end{abstract}


\section*{Introduction}
The animation, comic, and game industries have experienced significant economic growth, boasting expansive markets and widespread consumer engagement. This success is attributable to the considerable commercialization of creative content, the continuous expansion of global audience engagement, and the heightened demand for diverse and immersive entertainment experiences. With the rising popularity of 3D animation and gaming, designers and developers are continually seeking advanced 3D modeling techniques to create anime characters that can enhance the immersive experience for audiences \cite{2023_CVPR_panohead, 9982378, Abdal_2023_CVPR}. This demand has prompted innovation in 3D character modeling to meet the expectations of the ever-expanding and discerning consumer base.

\begin{figure*}[ht]
\centering
\includegraphics[width=\textwidth]{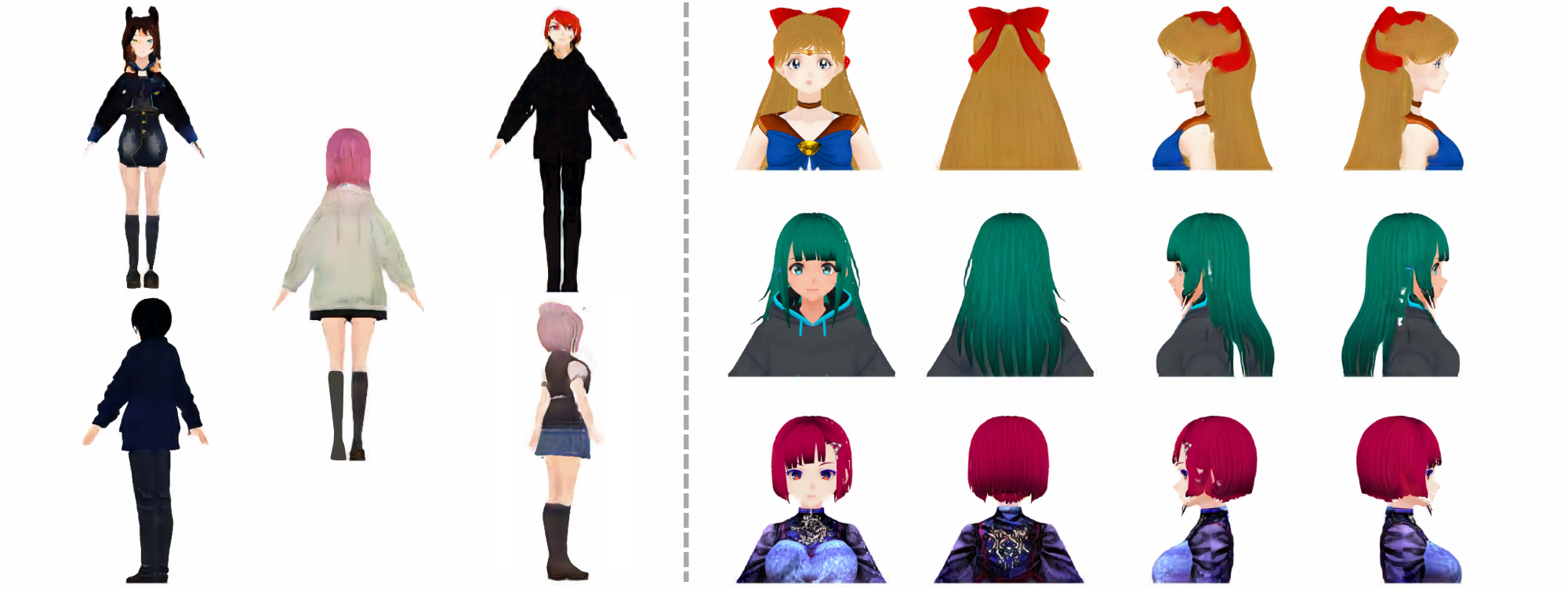} 
\caption{{\bf Left:} NOVA-3D achieves full-body anime character reconstruction from non-overlapped views. {\bf Right:} The results of NOVA-3D on head reconstruction of anime characters, with exquisite texture details, clear contours as well as 3D consistency.}
\label{fig:preface}
\end{figure*}

In the creation of anime characters, the conventional process in the industry begins with the design of a concept prototype. This concept prototype outlines the character's features, proportions, and design elements in a standardized way, providing a reference for the subsequent time-consuming 3D modeling stage. Designers typically start with front and back views of the character to ensure consistency and accuracy in guiding the production of the 3D models. The process of reconstructing geometry and applying texture mapping, reliant on 3D artists' expertise, is resource-intensive. Despite the pressing need for automation, the industry currently lacks full-body automatic generation solutions of 3D anime characters. One major challenge is the automated reconstruction lies in the scarcity of image sources for existing methods \cite{zhang2023large}. Our research aims to overcome the challenge of data scarcity by utilizing non-overlapping front and back view images of anime characters collected from industrial practice, and proposes an automated and efficient method for reconstructing 3D full-body models.

The task of reconstructing 3D avatars from images has been garnered significant interest in the field. However, the direct application of existing methods, such as neural radiance fields (NeRF) \cite{mildenhall2020nerf}, is limited in our proposed context. While showing potential in multi-view reconstruction, NeRF hinges on the availability of multiple and overlapped views. Such label-intensive resources are not routinely available in many industrial scenarios. Dual-view input adaptations of NeRF \cite{r16_4}, while technically feasible, tend to empirically produce reconstructions of 3D models that are relatively blurry and lack intricate details.

To address these challenges, we propose an innovative baseline method for 3D anime character modeling based on non-overlapped images. To better aggregate features from front and back views according to the observation direction, we introduce a direction-aware attention module for the fusion of front and back view features, enhancing 3D consistency. Another challenge is to capture intricate low-frequency details in the images, particularly from the back view. To preserve details in the generated back view images, we introduce a dual-viewpoint encoder module for extracting features at different granularities from both front and back views.

To facilitate this line of research, we meticulously curated the NOVA-Human dataset, comprising 10.2k anime models, multi-view images, and associated camera parameters. This dataset includes both random views and fixed orthogonal views, resulting in a total collection of 163.2k images. In summary, our contributions are as follows:

\textbullet \ We propose to investigate a new task of reconstructing 3D anime characters from non-overlapped views, as a more accessible and efficient solution for the industry.

\textbullet \ We introduce NOVA-3D, a novel baseline method to reconstruct 3D full-body characters from non-overlapped front and back view images. 

\textbullet \ Extensive experiments demonstrate that our method attains the best fidelity and quality compared with other state-of-the-art methods in 3D anime character reconstruction. Besides, our method achieves much higher quality when applied to the task of anime character head reconstruction \cite{r05}, which validates the effectiveness of our method.

\textbullet \ We establish a NOVA-Human dataset of 3D anime characters, encompassing 10.2k anime models with multi-view images and corresponding camera parameters. We will release the dataset to further nourish the research community. 
\begin{figure*}[t]

    \centering
    \includegraphics[width=\textwidth]{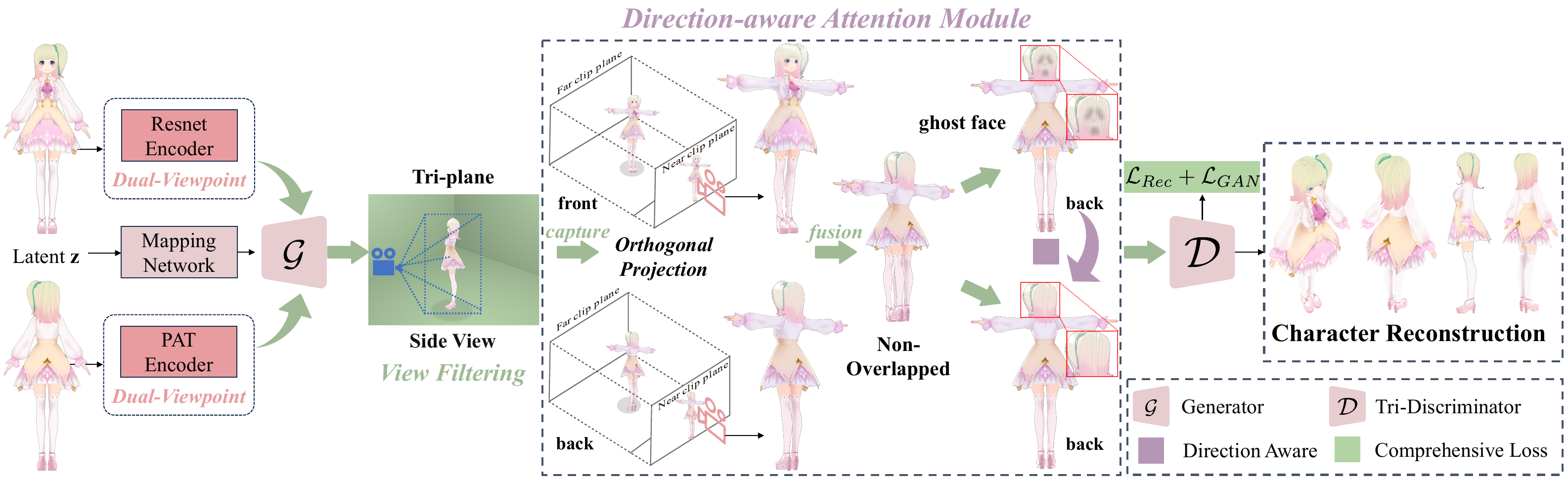}
    \caption{ The overall pipeline of NOVA-3D. NOVA-3D utilizes front and rear viewpoint images as input. The dual-viewpoint encoder extracts features from the images, which are then used by the generator to produce two tri-planes. The tri-planes are sampled to obtain sampling features, and the direction-aware attention module is employed to fuse the features. Finally, the reconstruction loss and GAN loss modules are used to calculate the overall loss.}
    \label{fig:pipeline}

\end{figure*}
\section{Related Work}
\subsection{3D generative model}
Recently, generative adversarial networks (GANs) \cite{r01} have managed to achieve great results in 2D image generation. The great success of GAN in the field of 2D image generation \cite{r02_1,r02_2,r02_3,r02_4,r02_5} has inspired many works to extend image generation from 2D to 3D \cite{r03_1,r03_2,r03_3,r03_4,r03_5,r03_6,r03_7,r03_8,r03_9}. These works achieve to learn the implicit representation of 3D scene information from a collection of 2D images, enabling generalized 3D reconstruction. EG3D \cite{r04} introduces the tri-plane representation, mapping information from 2D images to tri-plane, thereby enhancing the expressiveness of 3D scene representation. However, tri-plane representations exhibit entanglement between features, particularly when representing features with significant domain gap, as well as the occurrence of the "ghost face" problem when rendering the back view of the human body.

\subsection{Sparse-view 3D reconstruction}
Sparse-view 3D reconstruction has always been a challenging problem. Many studies \cite{r06_1,r06_2,r06_3,r06_4} have attempted to solve this problem. Among these, MVSNeRF \cite{r07} and GeoNeRF \cite{r08} are generalizable across different scenes. GeoNeRF demonstrates improved performance in inferring complex scenes with occlusion through the utilization of a multi-head attention mechanism and a cascaded cost volume for better geometric appearance. However, these methods need overlapped views for reconstruction, and at least three views are required as input. SparseNeRF \cite{r09} and DS-NeRF \cite{r10} enhance the performance of novel view synthesis from sparse-view by leveraging depth priors obtained from pre-trained deep models, high-precision depth scanners or multi-view stereo (MVS) algorithms. These depth priors are utilized to supervise NeRF in learning depth information. RegNeRF \cite{r12}, FreeNeRF \cite{r13} and InfoNeRF \cite{r14} have successively proposed regularization methods such as geometric regularization, appearance regularization, frequency regularization, and occlusion regularization to further improve the reality of 3D reconstruction. However, these methods based on either depth prior or regularization are unable to be directly applied to anime character reconstruction from non-overlapped image input.

\subsection{Avatar reconstruction}
With the emergence of the concepts of metaverse and virtual digital humans and the development of 3D reconstruction technology, more and more works are focused on avatar reconstruction \cite{r16_2,r16_3,r16_4,r16_5}. Knoll \textit{et al}. \cite{r17} proposed a method to create a novel surface alignment representation by surrounding a surface coordinate system around the human body, which can achieve high fidelity 3D avatar reconstruction. Most of these tasks are related to the reconstruction task of photo realistic humans, and they perform poorly in the domain of anime characters. Therefore, there have been some 3D human body reconstruction works applied in the domain of anime characters \cite{r05,r05_1}. PAniC-3D \cite{r05} proposed a method for reconstructing 3D head of an anime character from a portrait image. Though PAniC-3D has made groundbreaking progress in the field of 3D reconstruction of anime characters, it is limited to generating only the head portion. So far, there has been no existing work capable of achieving the 3D full-body reconstruction of anime characters. Therefore, the primary challenge of this task is to reconstruct 3D full-body anime characters with high controllability, filling the gap in the domain of 3D anime character reconstruction.

\section{Methodology}
Given the non-overlapped front and back views of an anime character, our target is to reconstruct a 3D full-body anime character. We formulate the reconstruction process as a 3D GAN model conditioned on non-overlapped images. We introduce a dual-viewpoint encoder that extracts features from front and back views with varying granularities (Section \ref{3.1}), a direction-aware attention module to aggregate features from two non-overlapped views (Section \ref{3.2}), and a comprehensive loss function (Section \ref{3.3}). The overall pipeline for our approach is illustrated in Figure \ref{fig:pipeline}.

\subsection{Dual-viewpoint encoder}
\label{3.1}
\begin{figure}[h]
\centering
\includegraphics[width=3in]{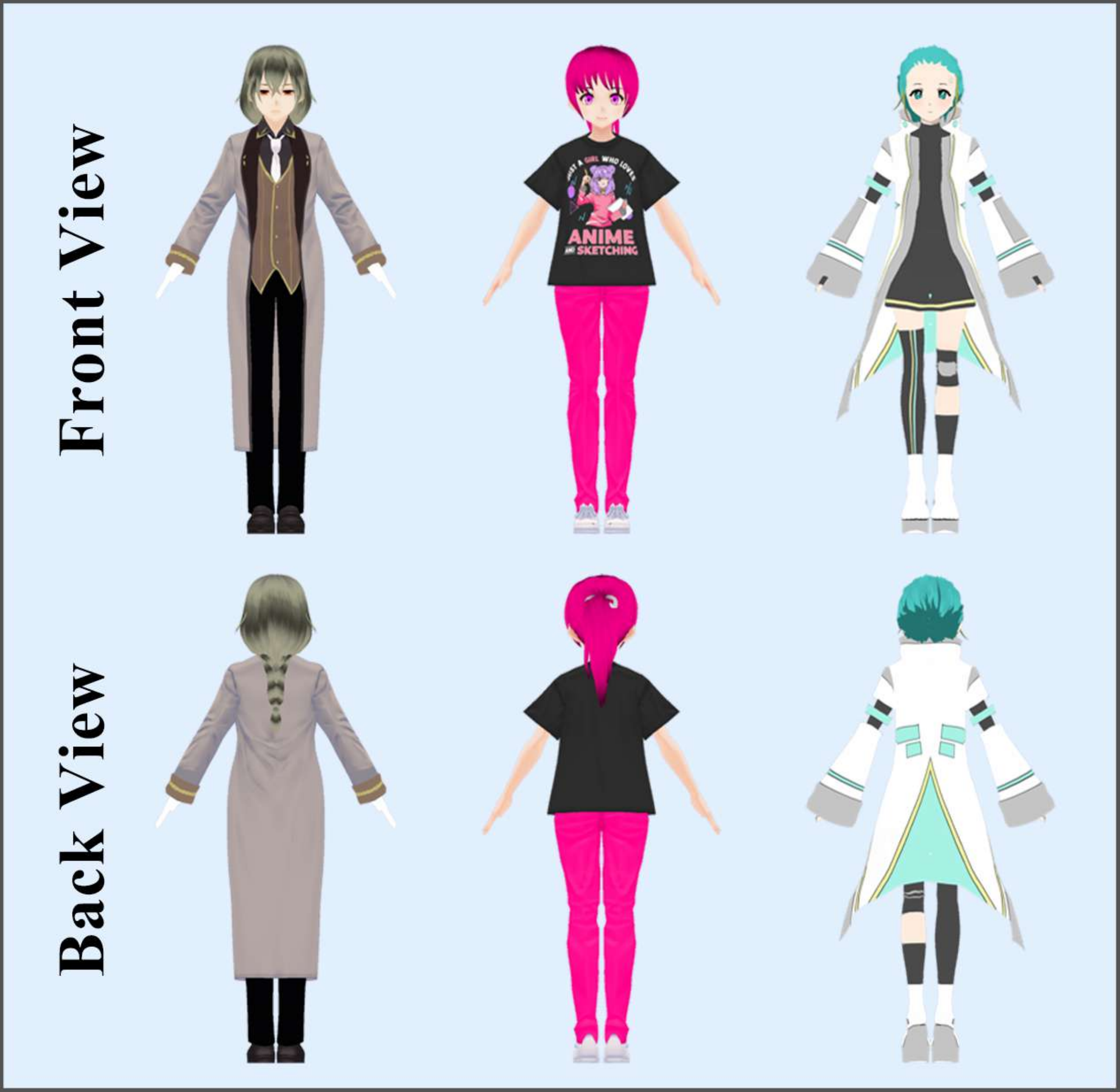}
\caption{Some examples of front and back images, where the front image contains more high-frequency information and the back image contains more low-frequency information.}
\label{fig:3-a}
\end{figure}

By employing neural networks to extract features from the front and back views, significant distinctions in feature granularity between the two views become evident. The front view of the anime character contains higher-frequency information, exhibiting richer textures and geometric shapes, such as facial features and frontal clothing. In contrast, the information encapsulated in the back view tends to be of lower frequency, presenting smoother textures for elements like monochromatic hair and the back of clothing, along with fewer geometric contours, as illustrated in Figure \ref{fig:3-a}. 

To address the disparity in high and low-frequency information between the front and back views, we propose a dual-viewpoint encoder, enabling the extraction of features adapted to the varying feature granularities. Higher-frequency information necessitates the use of a feature extraction module with a simpler structure, and vice versa. For the front view of the character body, owing to its rich texture, we employ ResNet for feature extraction. However, when the same feature extraction module is applied to the back view, compared to the front view, ResNet often lacks the capacity to capture low-frequency information adequately, resulting in a loss of detail and increased blurriness in the back view. Therefore, we switched to using the low-frequency information-sensitive PAT feature extraction module \cite{zheng2023potter} for feature extraction. Technically, to overcome the challenge of high-resolution degrading to low-resolution after patch merging, PAT introduces a high-resolution (HR) stream. This HR stream leverages local and global features from the base stream and employs patch splitting to preserve a high-resolution representation. This ensures the high-precision retention of both global and local features during the feature extraction process. Additionally, the pooling attention layer designed in PAT significantly reduces computational costs.

\subsection{Direction-aware attention module}
\label{3.2}
\begin{figure}[h]
\centering
\includegraphics[width=3in]{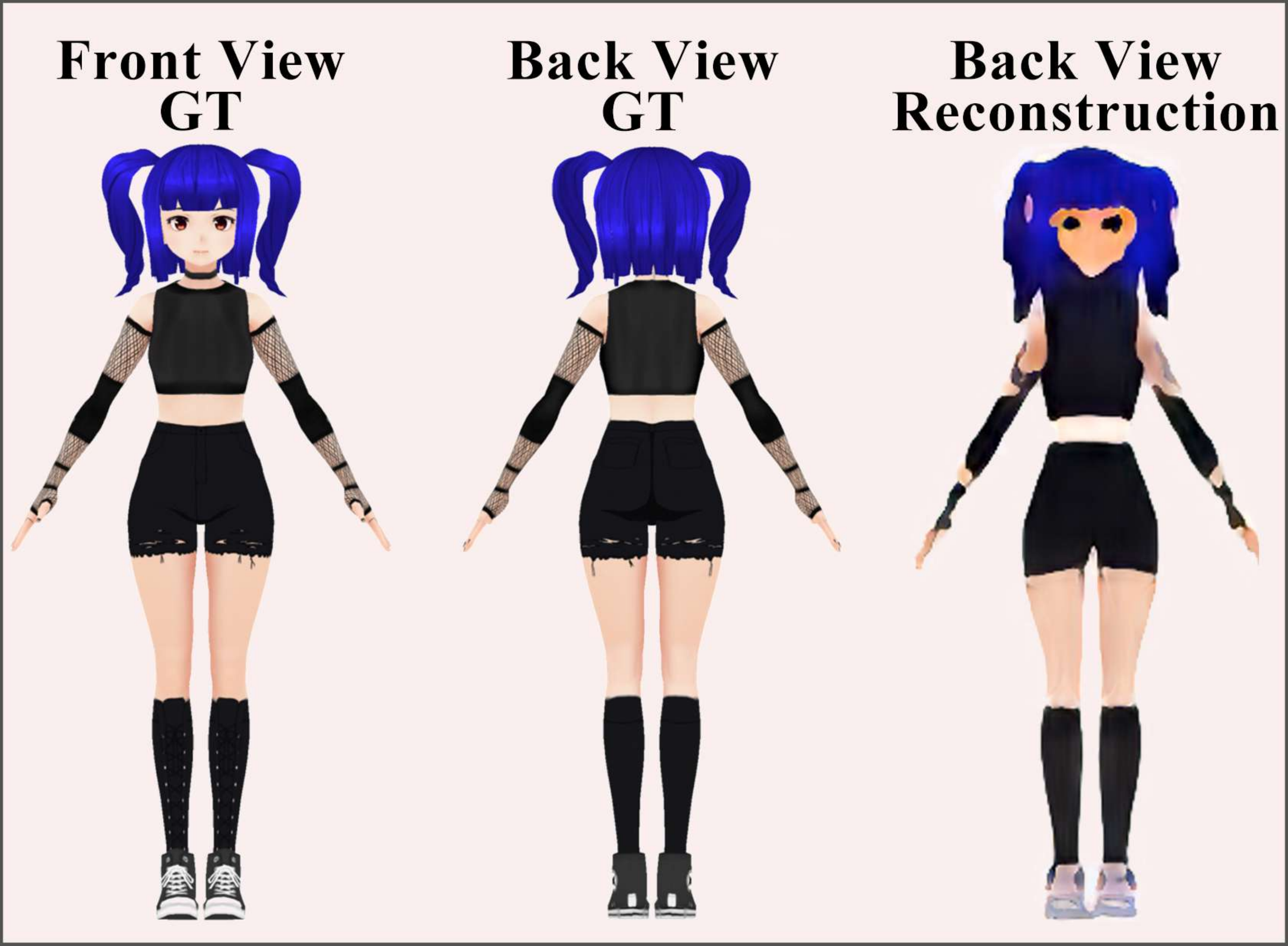}
\caption{Directly concatenating the features of the front and back view will lead to ghost face in the back view.}
\label{fig:3-b}
\end{figure}

We observed that simply using the front and back views to generate tri-plane representations and directly concatenating the extracted features together can lead to the emergence of frontal view textures when synthesizing the back view, resulting in issues such as ghost faces, particularly in the region of back hair, as shown in Figure \ref{fig:3-b}. Many recent studies \cite{r08,DEFN,Wonder3D} have utilized attention mechanisms to aggregate features extracted from multiple view images. These methods aim to integrate features extracted from different viewpoints to capture a holistic representation of the scene. However, challenges exist in effectively aligning features, handling inconsistent viewpoints, and maintaining spatial relationships, which may result in information loss and compromise feature integrity during the aggregation process. Motivated by the aforementioned works and to address this issue, we propose a direction-aware attention mechanism that better integrates features from the front and back perspectives, yielding more plausible back image synthesis.

To address the aforementioned issue, we propose a direction-aware attention module to facilitate the late fusion of features from two non-overlapped views. Leveraging the StyleGAN2 backbone \cite{karras2020analyzing}, we use the non-overlapped images of the front and back as conditions. In the construction of the camera system, we assume alignment of the world coordinate system origin with the object, with the face oriented towards the positive \textit{z}-axis. Based on the orthogonal projection, the front image aligned with the world coordinate system is used to generate a front-view tri-plane space, while the back image input generates a back-view tri-plane space, with the \textit{z}-axis and \textit{x}-axis opposite to those of the world coordinate system. Thus, when the camera in the world coordinate system takes a shot, the coordinates of the rays emitted by the camera need to be projected into the back-view tri-plane space, with the z-axis and \textit{x}-axis reversed. We sample \textit{N} points each from the front view tri-plane and the back-view tri-plane space, and extract features $\boldsymbol{F}_{front}, \boldsymbol{F}_{back}\in \mathbb{R}^{N\times 16}$.

We aim to prioritize tri-plane-extracted features from the front non-overlapped view when the randomly positioned camera is closer to the subject's front. Conversely, we rely more on back features when approaching the back. To achieve this, we introduce a direction-aware attention mechanism during the feature fusion process, allowing the feature fusion process to be modulated by the target view. Our attention module takes features from the front and back views as well as the observation direction as input, where \textit{K} and \textit{V} are obtained by concatenating the features from the front and back views. \textit{Q}, \textit{K}, \textit{V} are derived as:
\begin{equation}
    \boldsymbol{Q}=\phi_{q}(\boldsymbol{r}), \quad 
    \boldsymbol{K}=\phi_{k}(\boldsymbol{F}),\quad 
    \boldsymbol{V}=\phi_{v}(\boldsymbol{F}),
\end{equation}
where $\boldsymbol{r}=(r_x,r_y,r_z)\in\mathbb{R}^{N\times3}$ is the ray direction, $\phi_{q},\phi_{k},\phi_{v}$ are linear layers, $\boldsymbol{Q},\boldsymbol{K},\boldsymbol{V}\in \mathbb{R}^{N\times 32}$ , and $\boldsymbol{F}\in \mathbb{R}^{N\times 32}$ is the concatenation of features extracted from front and back views $\boldsymbol{F}_{front}, \boldsymbol{F}_{back}\in \mathbb{R}^{N\times 16}$.
% K is obtained as:
% \begin{equation}
%     K=\phi_{k}(concat(x_{front},x_{back})),
% \end{equation}
% where $x_{front}, x_{back}\in \mathbb{R}^{N\times 16}$ respectively represent the features obtained after passing the front and back views through the feature extraction module, and $K \in \mathbb{R}^{N\times 32}$ is a single-layer linear transformation. V is obtained as:
% \begin{equation}
%     V=\phi_{v}(concat(x_{front},x_{back})), V \in \mathbb{R}^{N \times 32},
% \end{equation}
To calculate the matching score, we have
\begin{equation}
    \boldsymbol{A}=\mathrm{softmax}(\frac{\boldsymbol{Q}\boldsymbol{K}^{T}}{\sqrt{d}}), \quad 
    % ,  A \in \mathbb{R}^{N*N}
    \boldsymbol{F}_{o}=\boldsymbol{W}_{l} \boldsymbol{A} \boldsymbol{V},
\end{equation}
% $\boldsymbol{F_o}$ is then obtained as:
% \begin{equation}
%      \boldsymbol{F_{o}}=\boldsymbol{W}_{l} \boldsymbol{A} \boldsymbol{V}
% \end{equation}
% \textbf{ F \in \mathbb{R}^{N \times 16}
where $\boldsymbol{W}_{l}$ is a learnable matrix used to generated final output feature $\boldsymbol{F}_{o}$.

According to our observations, the use of direction-aware attention effectively aggregates features from the front and back view images based on the observed viewpoint. This approach allows for a greater utilization of the features provided by the front image when closer to the front view, and vice versa, thus effectively resolving the ghost face issue.

Moreover, we employ tri-discriminator, utilizing upsampled RGB images, foreground masks for artifact removal from the background, and super-resolution images as supervisory signals. This approach leads to the achievement of high-quality 3D anime character reconstruction.

\subsection{Loss function}
\label{3.3}
\subsubsection{Reconstruction Loss}
We begin by inputting images from front and back views and render RGB image $I^r$, mask image $M$, and depth image $D$. After passing through a super-resolution network, we obtain RGB image $I^+$, and by bilinear interpolation, we derive super-resolved mask $M^+$ from \textit{M}. We calculate the reconstruction loss by comparing $I^+$, $M^+$, and $D$ with the ground truth RGB image $I_{gt}$, mask image $M_{gt}$, and depth image $D_{gt}$, respectively. The reconstruction loss is applied to four orthogonal viewpoints: front, back, left, and right, which are:

\begin{equation}
\begin{split}
     L_{rec} &= \lambda_{lpips} \text{LPIPS}(I^+, I_{gt}) \\
    &+\lambda_{l_1}|I^+-I_{gt}| \\
    &+\lambda_{mask}||M^+-M_{gt}||^2_2 \\
    &+\lambda_{depth}||D-D_{gt}||^2_2,   
\end{split}
\end{equation}
where $\lambda_{lpips}$, $\lambda_{l_{1}}$, $\lambda_{mask}$, and $\lambda_{depth}$ are hyperparameters used to balance the different loss components.

\subsubsection{Adversarial Loss.}
Only relying on the reconstruction loss can lead to model overfitting, hindering the synthesis of detailed novel-view images. Hence, we introduce an adversarial loss to ensure that images from randomly chosen viewpoints are more realistic. Specifically, we train the generator with front and back non-overlapped views $I_{front}$ and $I_{back}$ as inputs. The discriminator $D$ takes $I^{r+}, I^+, M^{+}$ as fake inputs and $I_{gt}$ and $M_{gt}$ as real inputs, where $I^{r+}$ is upsampled from $I^r$. The loss functions for training the generator and discriminator are as follows:
\begin{equation}
    L_{G} = -D(G(I_{front},I_{back})),
\end{equation}
\begin{equation}
    L_{D} = D(I^{r+},I^+,M^+) + (-D(I_{gt},I_{gt},M_{gt})),
\end{equation}
where \textit{D} represents the discriminator, while \textit{G} represents the generator.

To achieve training stability, we employ R1 regularization when training the discriminator.

\subsubsection{Regularization Loss.}
To ensure geometric continuity in space, we apply an L1 smoothness loss to the volume density:

\begin{equation}
    L_{reg} = \lambda_{reg}|\sigma(c)-\sigma(c+\epsilon)|,
\end{equation}
where $\sigma(\cdot)$ represents the volume density value corresponding to a spatial point, $c$ denotes the coordinates of a spatial point $c=(x,y,z)$, $\epsilon \in \mathcal{N}(0, 0.04^{2})$ represents a randomly applied perturbation, and $\lambda_{reg}$ is a hyperparameter indicating the weight of this loss.

\subsubsection{Overall Loss.}
The total loss is given by:
\begin{equation}
    L_{total} = L_{rec} + L_{G} + L_{D} + L_{reg}.
\end{equation}

More detailed implementation information and hyperparameters are given in appendix.

\section{NOVA-Human Dataset}
We curated a large dataset comprising 10.2k 3D anime character models from the open-source platform VroidHub. Our anime character models are constructed using an A pose, which in contrast to the traditional T pose, aligns more naturally with the typical postures of anime characters. Utilizing Vroid VRM models and employing ModernGL \cite{r36,r37}, we rendered multi-view images of the models. The images were rendered with unit ambient lighting, maintaining a distance of 3.5 units from the human body in VRMs \cite{r35}. To ensure consistent lighting effects and avoid inconsistencies in appearance between rendered images, we modeled only diffuse reflection lighting. The rendered images have a resolution of 512$\times$512.

Each anime model in the NOVA-Human dataset includes 16 randomly sampled views and 4 fixed orthogonal views, providing ample perspectives to address various model requirements for multi-view images. Each 3D model has 16 renderings based on randomly sampled views, where azimuth angles are sampled with a mean of 0 and a standard deviation of 100k, ensuring uniform sampling of azimuth angles. Elevation angles are sampled with a mean of 0 and a standard deviation of 20, resulting in most images having elevation angles close to 0. The fixed field of view was set at 30 degrees. In total, we generated 163.2k images, each accompanied by accurate camera parameters. Additionally, for each anime character model, we captured four images under orthogonal projection: front, back, left, and right views.

NOVA-Human dataset is of great significance for research and development in 3D anime character model reconstruction. Researchers and developers can leverage our dataset to train and evaluate their models in order to enhance the quality and accuracy of reconstructed anime character models.

\section{Experimentation}
In this section, we conducted a series of experiments to evaluate the reconstruction quality of NOVA-3D and demonstrate the effectiveness of the components in our architecture.

\subsection{Experimental settings}
\subsubsection{ Dataset}
Firstly, we utilized NOVA-Human dataset and conducted experiments with an image resolution of 1024$\times$1024. We selected three representative character models from the dataset and performed experiments on our method as well as baselines to assess the performance of our dataset. Then we use the Vroid 3D Dataset from PAniC-3D \cite{r05} to compare the results of PAniC-3D and NOVA-3D on the reconstruction of anime characters' heads.

\subsubsection{ Evaluation metrics}
We quantitatively evaluate the methods using Fréchet Inception Distance (FID) \cite{FID} which represents the variety and quality of the generated images.

\subsubsection{Implementation details}
NOVA-3D is implemented with PyTorch. In all experiments, we use a total batch size of 32 on 4 Nvidia A100 GPUs with fp16 precision. We utilized a pre-trained StyleGANv2 \cite{StyleGANv2} discriminator on the FFHQ dataset \cite{FHHQ}. Our optimization method is the Adam optimizer \cite{Adam}. The generator and discriminator have different learning rates, with the generator's rate being 0.0025 and the discriminator's rate being 0.0002. Further implementation details can be found in the appendix. We utilized L1 regularization to constrain the parameters of our model, with a regularization coefficient of 5.0 for the R1 regularization in StyleGANv2 and 1.0 for the regularization in our segmentation branch. For our reconstruction losses, we used a combination of LPIPS, L1, alphaL2, and depthL2, with equal weights assigned to each loss for different perspectives. The weights for LPIPS, L1, alphaL2, and depthL2 were 20.0, 4.0, 1.0, and 1000.0, respectively.

\subsection{Comparison results}

\subsubsection{ Baselines} We compared our method with two categories of baseline methods. 

\subsubsection{ Single-view reconstruction methods} 
We selected PAniC-3D for the reconstruction of single-view anime characters because it achieves the transition from illustrations to 3D half-body anime models. We compared our method with PAniC-3D to demonstrate that in the field of anime character reconstruction, we achieved finer and higher-quality reconstruction. 

\subsubsection{ Multi-view reconstruction methods}
For multi-view reconstruction, we compared our method (NOVA-3D) with MVSNeRF and GeoNeRF. Since NeRF-based methods require overlapping input images for reconstruction, MVSNeRF and GeoNeRF require at least three input images and can only perform reconstruction in limited angles near the input viewpoint. Therefore, we provided three viewpoints near the target viewpoint for reconstruction to MVSNeRF and GeoNeRF, comparing the reconstruction quality near the target viewpoint with our method. 

For a fair comparison, when testing on the Vroid dataset, we modified PAniC-3D by taking the front and back views as inputs, generating two tri-planes, and directly concatenating sampled features for training. This modified version is referred to as MV-PAniC-3D, and we compared the reconstruction quality of PAniC-3D, MV-PAniC-3D, and NOVA-3D.

\subsection{Results}

We compared NOVA-3D with baselines in two application scenarios: the reconstruction of the head of anime characters and the reconstruction of the entire body of anime characters.

\subsubsection{ Anime character head reconstruction}

\begin{figure}[h]
\centering
\includegraphics[width=3in]{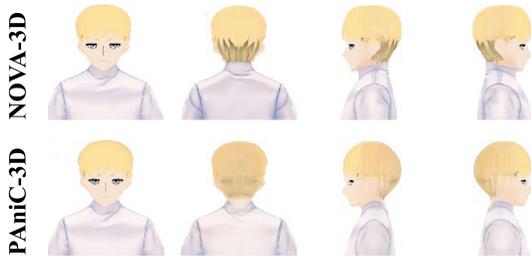}
\caption{{\bf Top:} The reconstruction results of NOVA-3D; {\bf Bottom:} The reconstruction results of PAniC-3D. NOVA-3D performs better than PAniC-3D in reconstruction on the same character.}
\label{fig:3d-a}
\end{figure}

\begin{figure}[h]
\centering
\includegraphics[width=3in]{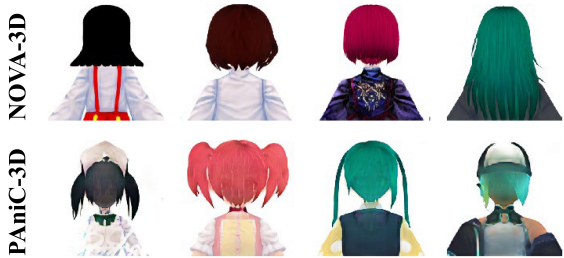}
 \caption{{\bf Top:} The reconstruction results of NOVA-3D; {\bf Bottom:} The reconstruction results of PAniC-3D. PAniC-3D has serious ghost faces in the back view, while our NOVA-3D method has completely removed ghost faces.}
\label{fig:3d-b}
\end{figure}

\begin{figure}[h]
\centering
\includegraphics[width=3in]{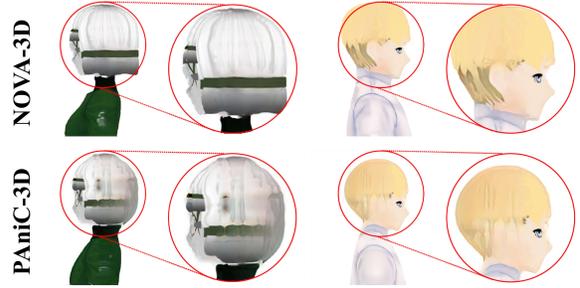}
\caption{Comparing the side views generated by NOVA-3D to those generated by PAnic-3D.NOVA-3D can generate novel views with richer details that are closer to the appearance of real characters.}
\label{fig:6-a}
\end{figure}

\begin{figure}[h]
\centering
\includegraphics[width=3in]{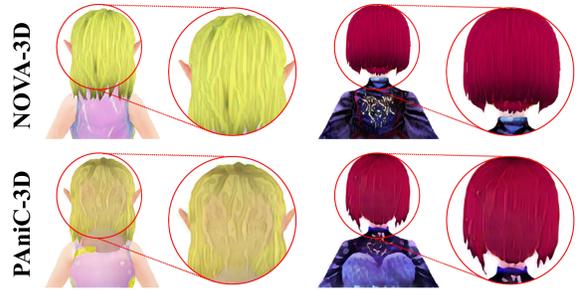}
\caption{Comparing the back views generated by NOVA-3D to those generated by PAnic-3D. NOVA-3D eliminates the symmetrical ghosting phenomenon.}
\label{fig:6-b}
\end{figure}

\begin{figure}[h]
\centering
\includegraphics[width=3in]{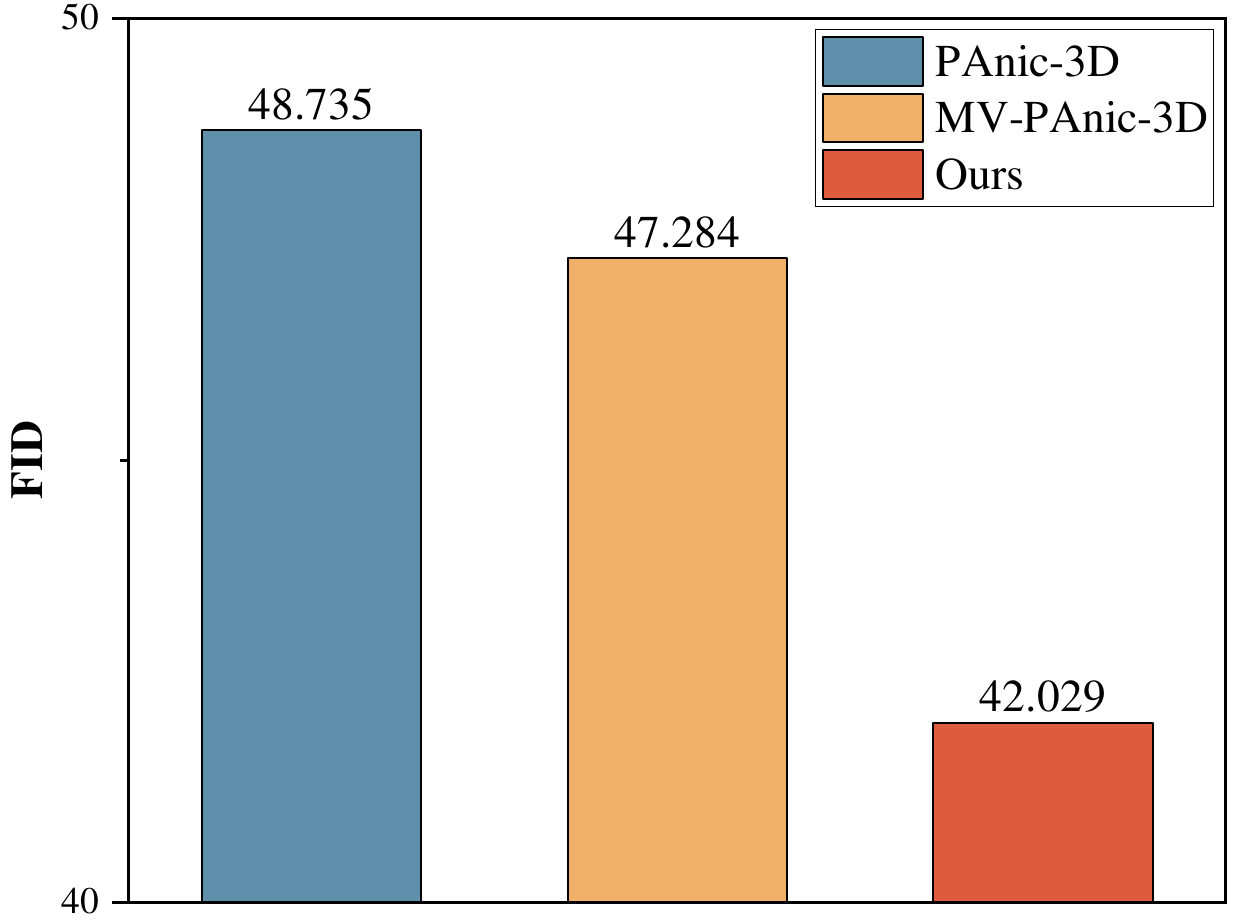}
\caption{Comparison of Fréchet Inception Distance for PAniC-3D, MV-PAniC-3D and NOVA-3D. Our proposed NOVA-3D framework performs better on the reconstruction of the same character than PAniC-3D and MV-PAniC-3D on the Vroid dataset.}
\label{fig:7-b}
\end{figure}

In the scenario of reconstructing the head of anime characters, to validate the effectiveness of our proposed method, we conducted experiments based on the Vroid dataset, comparing NOVA-3D, PAniC-3D, and MV-PAniC-3D. The experimental results are shown in Figure \ref{fig:3d-a}, Figure \ref{fig:3d-b}, Figure \ref{fig:6-a}, Figure \ref{fig:6-b}, and Figure \ref{fig:7-b}. Based on the experimental results, we performed both quantitative and qualitative analyses. Our method not only demonstrated significant visual advantages in discernible character features but also achieved a notable improvement in evaluation metrics.

\subsubsection{ Qualitative Analysis}
As shown in Figure \ref{fig:3-a} and Figure \ref{fig:3d-a}, our proposed NOVA-3D method excels in capturing high-frequency texture features of characters, accurately reconstructing fine details such as facial features and hair textures. For the same anime character, compared to PAniC-3D, our method incorporates a dual-viewpoint encoder. It utilizes different feature extraction modules for the front and back views, extracting features adapted to varying granularities. This results in a clearer rendering of high-frequency details, such as hair, ear contours, and clothing textures, as shown in Figure \ref{fig:3d-a}.

Furthermore, in the back view of the character, our method exhibits significantly superior reconstruction compared to PAniC-3D. When rendering the back view, PAniC-3D suffers from a "ghost face" problem due to the entanglement of features from the front and back views, as shown in Figure \ref{fig:3d-a} and Figure \ref{fig:3d-b}. It significantly compromises the rendering quality of the character's back view in PAniC-3D. \textbf{NOVA-3D effectively addresses the "ghost face" issue by employing the direction-aware attention module proposed in our approach.} This module separately constructs tri-plane spaces for features from the front and back views and performs subsequent fusion. The result is a realistic and clear rendering of the character's back view, as shown in Figure \ref{fig:3d-a}. These qualitative results indicate that our method surpasses the current state-of-the-art method, PAniC-3D, in terms of reconstruction quality and detail fidelity.

\subsubsection{ Quantitative Analysis}
To assess the ability to capture the intricate details which commonly present in anime characters, we employ the Frechet Inception Distance (FID) metric which accurately measures the effectiveness of generating high-frequency details and provides a more comprehensive evaluation of generative performance. 

In this section, we compare PAniC-3D, MV-PAniC-3D, and our method using FID as the metric. As shown in Figure \ref{fig:7-b}, \textbf{our NOVA-3D outperforms existing state-of-the-art (SOTA) methods in terms of FID.} PAniC-3D uses a single view as input, leading to the loss of features on the opposite side. In comparison, MV-PAniC-3D, which uses dual viewpoints as input, for lack of a direction-aware attention mechanism to fully utilize the features of front and back viewpoints results in difficulties selecting features related to the input direction without overlap, causing feature coupling in the reconstructed character views. \textbf{Our method demonstrates superior performance in terms of FID, achieving a reduction of 5.255 compared to MV-PAnic-3D and a reduction of 6.706 compared to PAnic-3D. }This indicates that our proposed NOVA-3D surpasses existing SOTA methods in both image quality and diversity.

\subsubsection{ Anime character full-body reconstruction}

Our method surpasses the current state-of-the-art methods in the scenario of reconstructing the head of anime characters, with reconstruction results far superior to the baselines. Moreover, we extended the application of NOVA-3D to the more challenging scenario of reconstructing the entire body of anime characters. We compared the reconstruction results with GeoNeRF, MVSNeRF, and PAniC-3D, as shown in Figure \ref{11111}.

\subsubsection{ Qualitative Analysis}
\begin{figure*}[t]
    \centering
    \includegraphics[width=0.9\textwidth]{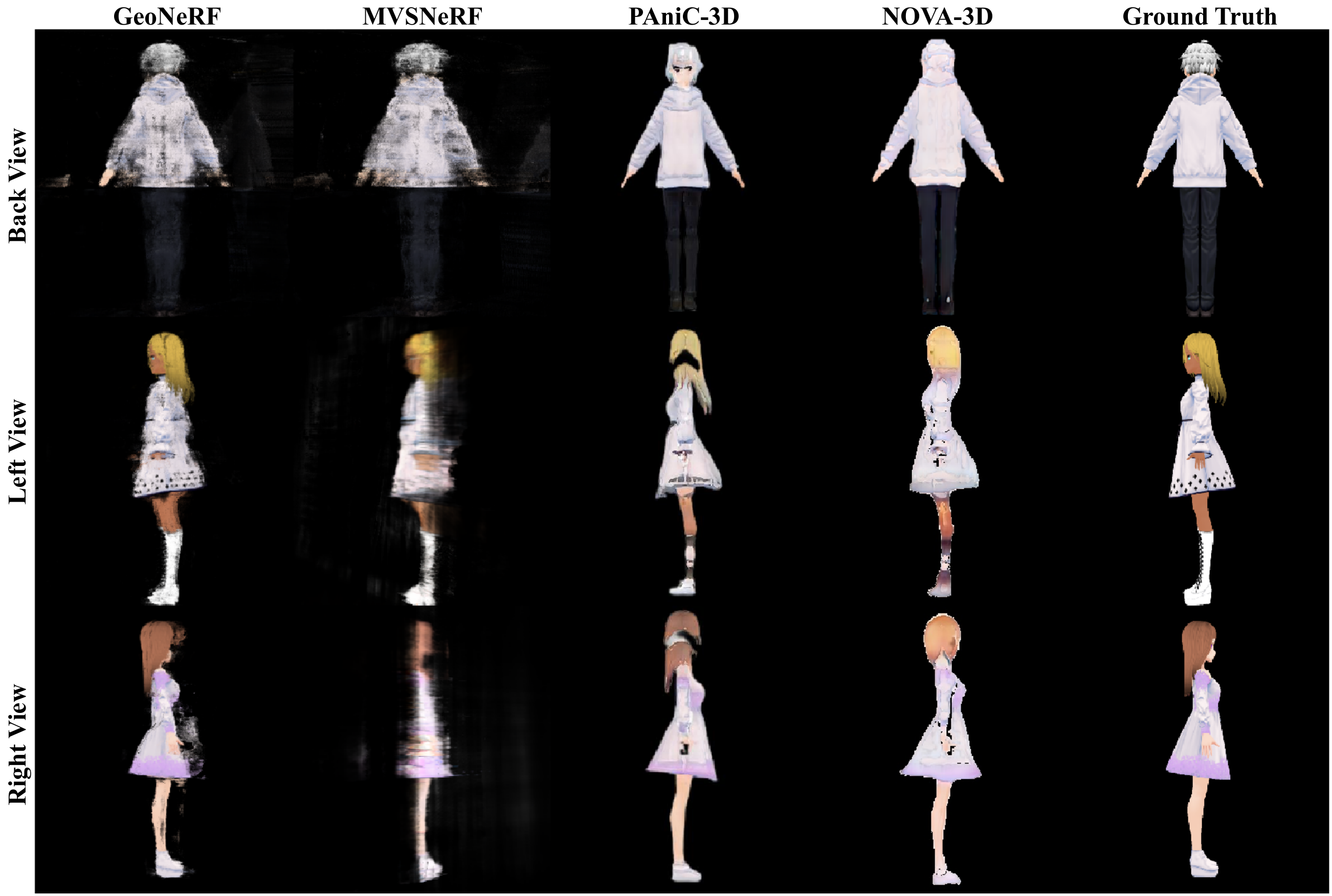}
    \caption{{3D reconstruction visualization. }We visualize the 3D reconstruction results across a wide range of methods on the NOVA-Human dataset. NOVA-3D outperforms the baselines by reconstructing a more complete and realistic anime human body.}
    \label{11111}
 
\end{figure*}

Based on the comparison experiments and the obtained full-body reconstruction results, we thoroughly considered several details of the character reconstruction and analyzed the reconstruction effects of both the left and right side views as well as the back view, as shown in Figure \ref{11111}. In the case of using only two front and back images as input, GeoNeRF and MVSNeRF, limited by NeRF's poor capability with sparse inputs, could only reconstruct very blurry images, accompanied by severe artifacts. In the left and right side views, they were unable to generate a complete outline of the human body. PAniC-3D, which uses the more powerful StyleGANv2 network to generate new views, was able to reconstruct a human body with clear outlines. However, it suffered from the problem of "ghost face" in the head region of the back view, which made the back view visually similar to the front view. In addition, the left and right side views reconstructed by PAniC-3D were affected by artifacts, resulting in the loss of some information and holes in the head region, so that the entire body could not be reconstructed.

In comparison to the baselines, our proposed NOVA-3D uses a generative approach to synthesize new views, achieving a relatively complete and clear reconstruction of the human body. In addition, we completely removed artifacts and holes in the background by using foreground masks. \textbf{The dual-viewpoint encoder and direction-aware module enhanced the quality of image details and textures, suppressing the generation of "ghosting face".} In conclusion, NOVA-3D achieves high-fidelity 3D full-body reconstruction of anime characters, surpassing the current state-of-the-art methods.

\renewcommand{\arraystretch}{2} %控制行高  
\begin{table*}[t]  
\caption{Ablation experiment on NOVA-Human datasets.}  
	
	\centering  
	\fontsize{8.5}{5}\selectfont  
		\begin{tabular}{ccccccccccccc}  
			\toprule  
			\multirow{2}{*}{Method}&\multicolumn{3}{c}{ Right}&\multicolumn{3}{c}{ Left}&\multicolumn{3}{c}{ Front}&\multicolumn{3}{c}{ Back}\cr  
			\cmidrule(lr){2-4} \cmidrule(lr){5-7}  \cmidrule(lr){8-10} \cmidrule(lr){11-13} 
			&SSIM↑&LPIPS↓&PSNR↑&SSIM↑&LPIPS↓&PSNR↑&SSIM↑&LPIPS↓&PSNR↑&SSIM↑&LPIPS↓&PSNR↑\cr  
			\midrule  
			PAniC-3D&94.710&7.475&19.685&94.817&{\bf 7.099}&{\bf 19.948}&93.814&7.637&19.725&92.699&{\bf 8.748}&{\bf 19.031}\cr  
			w/o DAM&93.722&10.964&17.643&93.684&11.233&17.534&90.743&13.201&16.780&90.818&13.615&16.321\cr  
			w/o DVE&94.728&8.578&19.167&94.757&8.266&19.227&93.791&8.593&19.442&92.713&9.977&18.377\cr  
			NOVA-3D&{\bf 95.114}&{\bf 7.333}&{\bf 19.816}&{\bf 95.155}&7.191&19.829&{\bf 94.248}&{\bf 7.627}&{\bf 19.677}&{\bf 93.293}&8.756&18.979\cr  
			\bottomrule  
		\end{tabular}

		\label{tab:performance_comparison}  
\end{table*}

\subsection{Ablation studies}
In this section, we conducted multiple ablation experiments on the NOVA-Human dataset and the Vroid 3D dataset to demonstrate the effectiveness of the proposed modules. For the ablation experiments, we continued to use Structural Similarity Index (SSIM), Learned Perceptual Image Patch Similarity (LPIPS), and Peak Signal-to-Noise Ratio (PSNR) as the evaluation metrics for the reconstruction results.

\subsubsection{Ablation of Direction-aware Attention Module (DAM)}
To compare the view-aware capacity of DAM in learning the features from two tri-plane samples, we conducted a comparative experiment. It is validated in Table \ref{fig:3d-b} and Figure \ref{11111}, where we observe that with the help of DAM, NOVA-3D can recover the texture details of character's face and clothing in both views. But when we remove this module results in the entanglement of features from the front and back views. This entanglement leads to the synthesis of back images that contain the frontal information ("ghost face"). As shown in Figure \ref{tab:performance_comparison}, when DAM is not incorporated, compared with NOVA-3D, the performance indicators of SSIM, LPIPS, and PSNR have all shown significant reductions, quantitatively demonstrating the effectiveness of DAM.

\subsubsection{Ablation of Dual-viewpoint Encoder (DVE)}
\begin{figure}[h]
\centering
\includegraphics[width=3in]{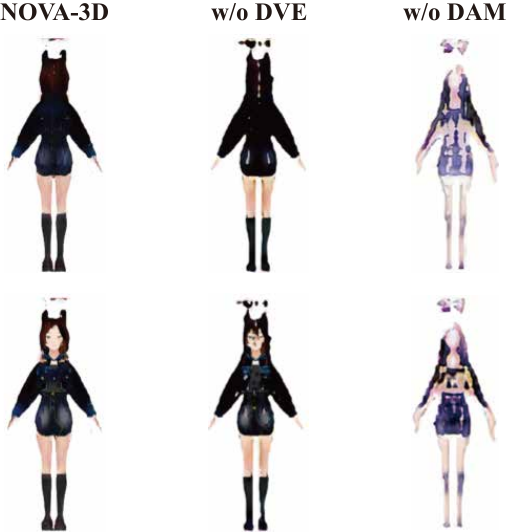}
\caption{Ablation study on visual effect. Three columns denote "w Dual-viewpoint Encoder and Direction-aware Attention Module" (NOVA-3D, ours), "w/o Dual-viewpoint Encoder" (DVE) and "w/o Direction-aware Attention Module" (DAM), respectively.}
\label{fig:7-a}
\end{figure}

Similarly, to validate the ability of the DVE module in extracting high-frequency texture details from images, we conducted comparative experiments with the DVE module. The DVE utilizes different feature extraction modules for the front and back views to adapt to their granularity and effectively extract features (\textit{i.e.}, facial features, and clothing textures) effectively. As shown in Figure \ref{fig:7-a}, when we remove this module and use the same ResNet encoder for feature extraction, the same ResNet encoder fails to effectively extract both high-frequency and low-frequency features at the same time (Figure \ref{fig:7-a}). This results in the loss of some high-frequency texture features, which generate white noise in the same parts of the rendered images, and a significant degradation of detail quality. As shown in Table \ref{tab:performance_comparison}, when DVM is not incorporated, compared with NOVA-3D, the performance indicators SSIM, LPIPS and PSNR have all shown reductions, providing quantitative evidence of the effectiveness of the DVE method.

\section{Conclusion and Future Work}
To address the current challenges in 3D full-body reconstruction of anime characters, we propose NOVA-3D, a framework for reconstructing 3D anime characters based on non-overlapped views. Through the collaborative integration of our designed dual-viewpoint encoder module and direction-aware attention module, we achieve the task of reconstruction based on non-overlapped views. Since the task involves a novel domain lacking an established dataset, we introduce the innovative NOVA-Human dataset containing 10.2k 3D anime characters with multi-view images and accurate camera parameters. This dataset can be utilized for research in the field of 3D anime character generation.

NOVA-3D, outperforms current state-of-the-art 3D reconstruction methods on the NOVA-Human dataset, successfully addressing the challenging task of reconstructing 3D anime character models from non-overlapped views. Our work meets the industry's demand for rapid and automatic reconstruction of 3D anime character models from non-overlapped views, alleviating the time-consuming and labor-intensive tasks faced by industry professionals who use orthogonal designs and specialized modeling software for 3D anime character modeling. This significantly lowers the barrier to entry for 3D anime modeling and enhances the productivity. However, our work has limitations, as it still relies on front and back non-overlapped views as inputs. Considering the current capabilities of large-scale diffusion models from text to images, we aim to explore the use of text or front views as prompts in the future, making our method more practical and convenient.

\label{sec:conclusion}

\bibliographystyle{ieeenat_fullname}
\bibliography{main}

\clearpage
\newpage

\appendix
\vspace*{1em}{\centering\Large\bf%
Appendix
\vspace*{1.5em}}

We curated a large dataset comprising 10.2k 3D anime character models from the open-source platform VroidHub. We named it the "NOVA-Human" dataset. The dataset has the following characteristics:

\textbullet \ Previously, there were hardly any large-scale datasets available specifically for full-body 3D anime character models.

\textbullet \ Instead of using the traditional T-pose for character model construction, we opted for the A-pose, which aligns more naturally with the typical poses of anime characters.

\textbullet \ Each character model in NOVA-Human has diverse viewpoints and precise camera parameters, which were rarely seen in the past.

We will analyze the superiority and practicality of this dataset from three different perspectives.

\begin{figure}[h]
    \centering
    \includegraphics[width=0.45\textwidth]{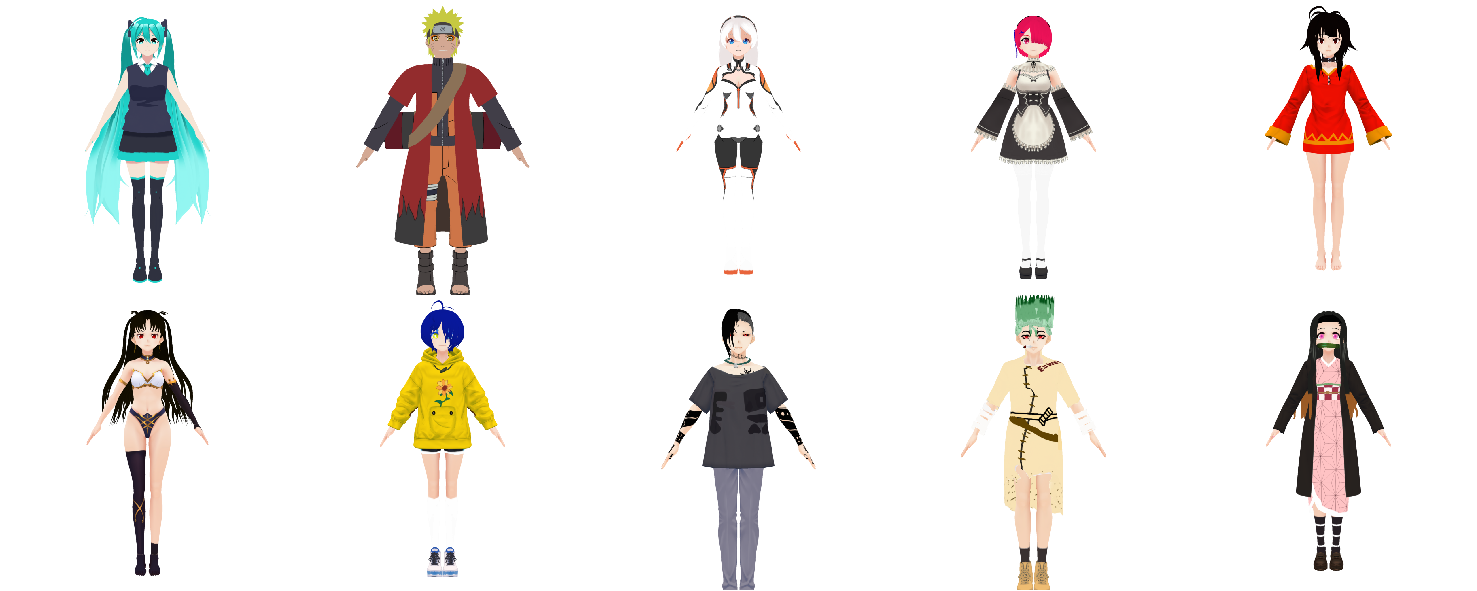}
    \caption{Here are some anime characters from various works in the ACG (Anime, Comic and Game) field showcased in the NOVA-Human dataset. It includes well-known characters in the ACG community such as Hatsune Miku, Uzumaki Naruto, Kiana, Ram, Megumin, and Kamado Nezuko.}
    \label{ACG}
\end{figure}

\section{Diversity}
\begin{figure}
\centering
\begin{subfigure}{1in}
  \centering
  \includegraphics[width=1in]{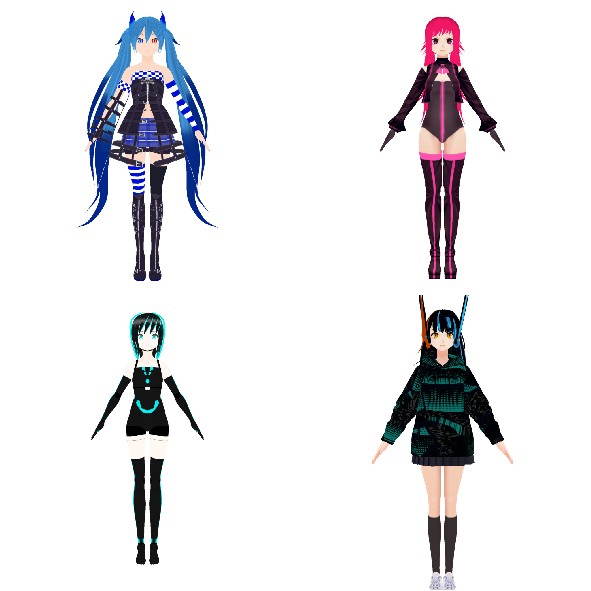}
  \caption{}
\end{subfigure}
\begin{subfigure}{1in}
  \centering
  \includegraphics[width=1in]{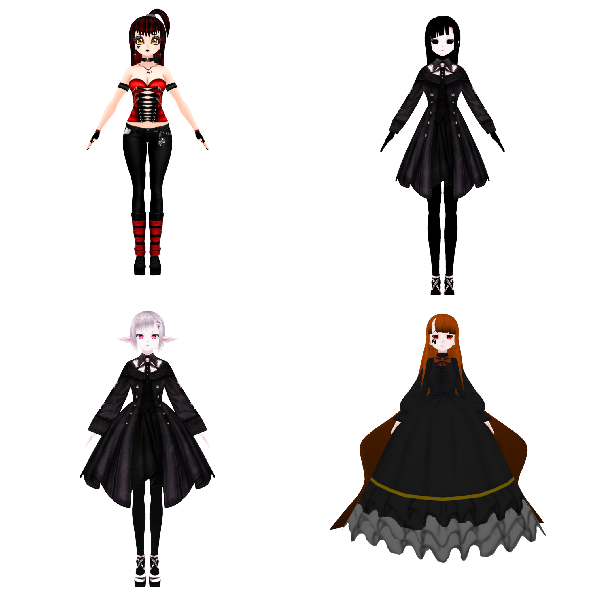}
  \caption{}
\end{subfigure}
\begin{subfigure}{1in}
  \centering
  \includegraphics[width=1in]{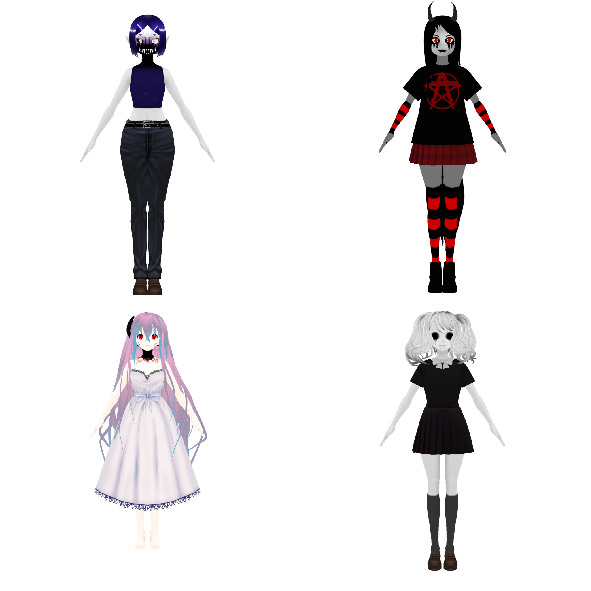}
  \caption{}
\end{subfigure}
\caption{Animated characters with various styles in the NOVA-Human dataset. (a) Futuristic or near-futuristic Cyberpunk style. (b) Gothic style. (c) Dark Horror style.}
\label{fig:three-styles}
\end{figure}

\begin{figure}[h]
    \centering
    \includegraphics[width=0.45\textwidth]{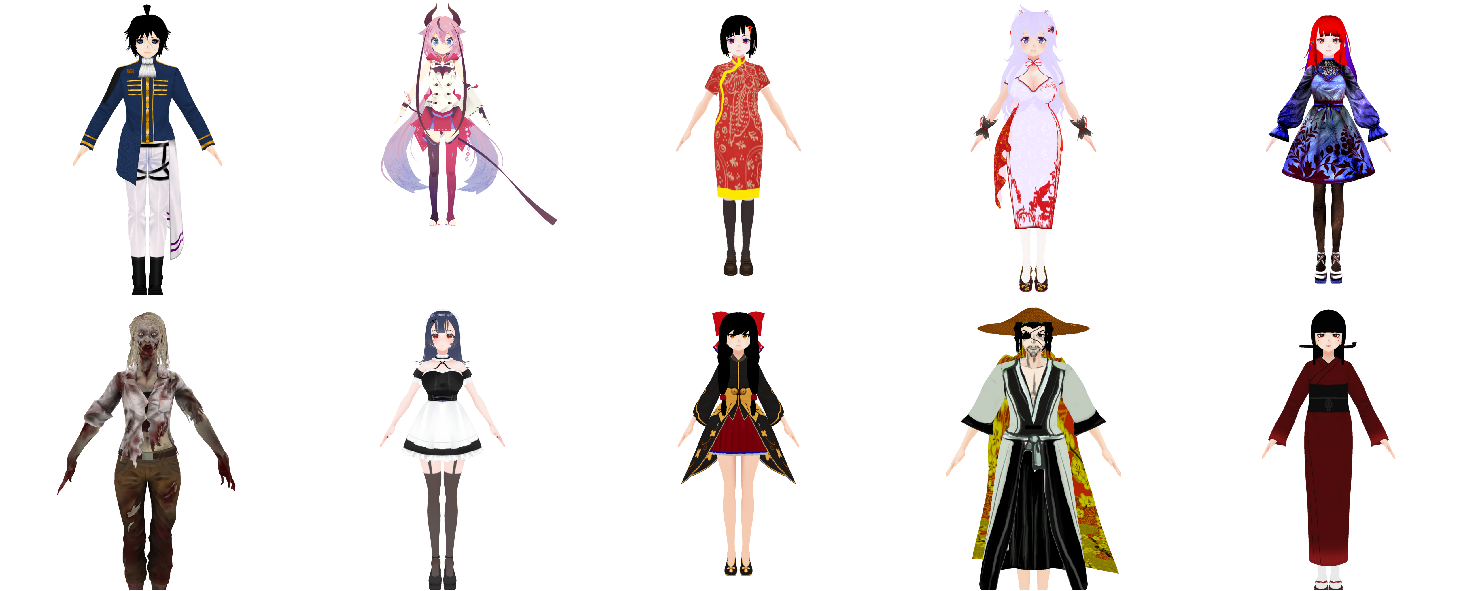}
    \caption{Here we showcase additional sampled views from the NOVA-Human dataset. Our dataset encompasses a diverse range of 3D anime characters.}
    \label{moredata}
\end{figure}

\begin{figure}[h]
    \centering
    \includegraphics[width=0.45\textwidth]{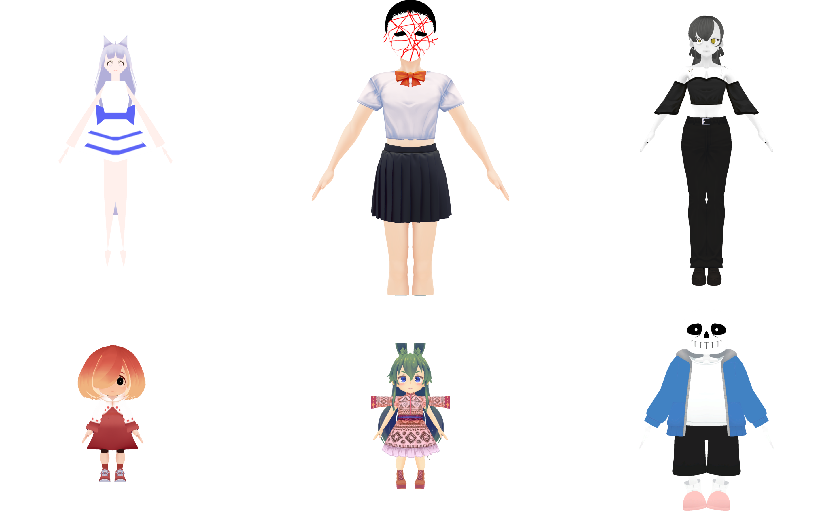}
    \caption{Here are anime characters with niche styles showcased in the NOVA-Human dataset, including various unique and unconventional characters.}
    \label{xiaozhong}
\end{figure}

To meet the requirements of multi-view 3D reconstruction, we provided a rich variety of viewpoints during the construction of the NOVA-HUMAN dataset. Each anime character model includes 16 randomly sampled viewpoints and 4 fixed orthogonal viewpoints. The randomly sampled viewpoints use perspective projection, while the orthogonal viewpoints use orthographic projection. The diversity of anime characters in the dataset is also one of its strengths. Previous research mostly focused on specific types of anime characters, whereas our dataset contains various styles of anime characters. It includes characters from popular Japanese anime and games, as well as many virtual anime creations from individual artists on VroidHub. The dataset encompasses a wide range of styles, such as traditional Japanese anime style (Figure \ref{ACG}), Cyberpunk and Futuristic style (Figure \ref{fig:three-styles}a), Gothic style (Figure \ref{fig:three-styles}b), Dark Horror style (Figure \ref{fig:three-styles}c), and various anime characters with distinct temporal and regional characteristics. The dataset also exhibits a plethora of clothing and accessories variations, as shown in Figure \ref{moredata}. Additionally, our dataset encompasses anime characters of different ages, body types, and races (\textit{e.g.}, Beast ear girl, Elf, Demons, Robots, etc.). Not only does it cover mainstream anime character styles, but it also includes numerous characters with unique and unconventional outfits, as shown in Figure \ref{xiaozhong}. The diversity of characters in our dataset ensures that users can find the models that they want to use.

\begin{figure}[h]
    \centering
    \includegraphics[width=0.45\textwidth]{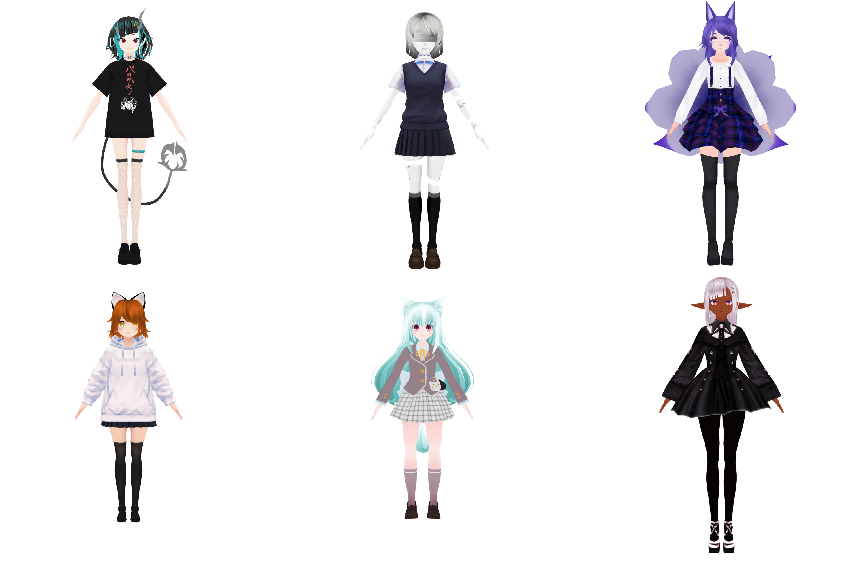}
    \caption{Here are the anime characters with different body types and races showcased in the NOVA-Human dataset, including Beast ear girl, Elf, Demons, Robots, and more.}
    \label{zhongzu}
\end{figure}

\section{Challenges} During the construction of the dataset, we encountered some challenges that were eventually overcome. By utilizing Vroid VRM models and ModernGL~\cite{r36,r37}, we rendered multi-view images of the models. These images were rendered using unit environmental lighting, maintaining a distance of 3.5 units from the human body in VRM~\cite{r35}. One of the difficulties in the data preparation process was determining the appropriate distance. After numerous attempts with different distances, we finally settled on using a distance of 3.5 units in VRM to capture complete and properly proportioned human figures. Furthermore, different lighting effects could lead to inconsistent appearances between rendered images. To address this issue, we only simulated diffuse lighting. The rendered images have a resolution of 512 × 512.

\section{Complexity}

\begin{figure}[h]
    \centering
    \includegraphics[width=0.45\textwidth]{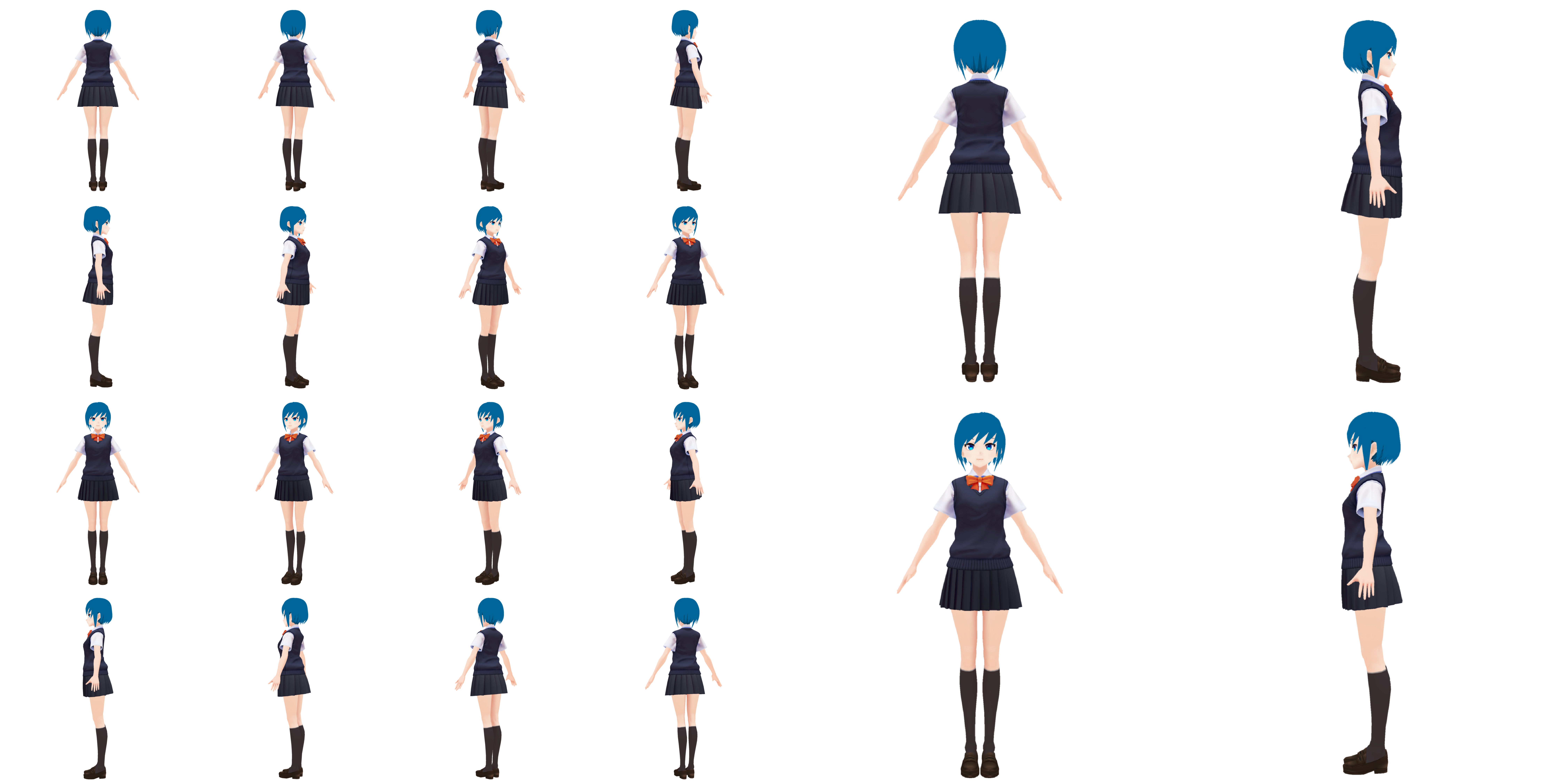}
    \caption{Here we present the training data sampled from a 3D anime character model in NOVA-Human. On the left are 16 randomly sampled views, while on the right are four fixed orthogonal sampled views.}
    \label{dataset1}
\end{figure}

For each model, we performed 16 renders based on randomly sampled viewpoints. During this process, the azimuth sampling used a mean of 0 and a standard deviation of 100k to ensure uniform sampling of azimuth angles. As for the elevation angle, the sampling mean was 0 with a standard deviation of 20, resulting in most images having an elevation angle close to 0. We chose a fixed field of view of 30 degrees, which is commonly used in standard lenses. In total, we generated 163.2k images with accurate camera parameters providing valuable information for further analysis and evaluation of models. Additionally, for each anime character model, we captured images from four orthogonal viewpoints: front, back, left, and right, using orthographic projection (Figure \ref{dataset1}).

The NOVA-Human dataset is vital for research and industry in 3D anime character model reconstruction.  Researchers and industrial developers can utilize our dataset to train and evaluate their models, aiming to improve the quality and accuracy of reconstructing anime character models.

\end{CJK}
\end{document}